\pdfoutput=1

\documentclass[11pt]{article}

\usepackage{EMNLP2023}
\usepackage{times}
\usepackage{latexsym}
\usepackage{xspace}
\usepackage[T1]{fontenc}

\usepackage[utf8]{inputenc}

\usepackage{microtype}

\usepackage{inconsolata}
\usepackage{balance}

\newcommand{\sys}{NeuSTIP\xspace}

\title{\sys: A Novel Neuro-Symbolic Model for Link and Time Prediction  \\ in Temporal Knowledge Graphs}

\author{Ishaan Singh \hskip 1em  
Navdeep Kaur  \hskip 1em
Garima Gaur \hskip 1em Mausam \\
  Indian Institute of Technology, Delhi \\
  \texttt{\{ishaanyuvraj, navdeepkjohal, garimagaur27\}@gmail.com} \\
   \texttt{mausam@cse.iitd.ac.in}
}
\begin{document}
\maketitle
\begin{abstract}
While Knowledge Graph Completion (KGC) on static facts is a matured field, Temporal Knowledge Graph Completion (TKGC), that incorporates validity time into static facts is still in its nascent stage. The KGC methods fall into multiple categories including embedding-based, rule-based, GNN-based, pretrained Language Model based approaches. However, such dimensions have not been explored in TKG. To that end, we propose a novel temporal neuro-symbolic model, \sys, that performs link prediction and time interval prediction in a TKG. \sys learns temporal rules in the presence of the Allen predicates that ensure the temporal consistency between neighboring predicates in a given rule. We further design a unique scoring function that evaluates the confidence of the candidate answers while  performing link prediction and time interval prediction by utilizing the learned rules. Our empirical evaluation on two time interval based TKGC datasets suggests that our model outperforms state-of-the-art models for both link prediction and the time interval prediction task.

\end{abstract}

\section{Introduction}

Knowledge Graphs(KGs) are the popular factual information repositories where each fact is encoded as a \emph{triple} ($ s,r,o$), where $s$ and $o$ be the real-world entities and $r$ be the relationship between them, for instance, the triple ($\textit{Joe Biden, presidentOf, USA}$) is representing the fact that \emph{Joe Biden} is the president of \emph{USA}. Interestingly, many of the entity-entity relations have temporal facet, for eg, $(\textit{Tim\_Berners\_Lee, wasBornIn, London, 1995} ),$ $(\textit{Einstien, workedAt, ETH\_Zurich, [1912,1914]}$), and so on. The modeling of the temporal aspect of facts leads to the collection of quadruples $(s,r,o, T)$ that are referred to as temporal KGs (TKGs). 
While being a popular source of structured information, KGs are often incomplete. To this end, the problem of enriching KGs by inferring missing information is formulated as a KG completion (KGC) task. In the context of static (non-temporal) KGs, the KGC involves \emph{link prediction} task, viz, given a query ($s,r, ?$), predict $o$ for which the fact ($s, r, o$) holds in the real-world. 
This problem is fairly well-studied and has been tackled from various standpoints. Existing works can be categorized based on their learning task formulation -- GNN-based solutions \cite{NBFNet2021, CompGCN2020}, LM-based approaches \cite{KGBERT2019, SimKGC2022}, KG embedding-based models~\cite{Rotate2019, complex2016}, and neuro-symbolic solutions \cite{NeuralLP2017, RNNLogic2021}.  



The problem of KGC becomes more challenging when the temporal aspect of facts is taken into consideration. The temporal knowledge graph completion (TKGC) involves an additional task of \emph{time} prediction, i.e. given a fact $(s,r,o,?)$, infer its temporal information. The majority of existing solutions \cite{HYTE2018,Timeplex2020, BoxTE2022} focus on learning time-aware latent representation of entities and relations. While the embedding-based methods have showcased reasonable performance for link prediction, the interpretability of such solutions is limited. Recent works TLogic~\cite{TLogic2021}, ALRE-IR~\cite{ALRE-IR2022}, and TILP ~\cite{TILP2023} are notable neuro-symbolic solutions that use interpretable temporal rules for inferring missing links in the temporal KG. The domain of TLogic and ALRE-IR is limited as they handle only facts with time instants. TILP proposes a decent solution for learning logical rules for time interval data but the framework does not address the time interval prediction. Interval prediction is an intrinsically hard problem. One of the challenges is to design a unified  rule language that is as intuitive and interpretable as the closed path non-temporal logical rules used in rule-based KGC approaches. Further, in existing works \cite{TLogic2021,TILP2023,RNNLogic2021}, the rule \emph{confidence} is derived solely from either the statistical measures or the similarity scores of latent representations. Therefore, no single technique leverages the goodness of explicit and implicit pattern information present in the TKG.


In response, our proposed work makes the following contributions:

\begin{itemize}
    \item We design an intuitive rule language that integrates the \emph{complete} set of Allen algebra relations and KG relations such that the link prediction and time prediction are performed from the vantage point of the principal entity present in a given query. Further, we propose an innovative way of computing time-aware rule confidence. 

    \item This work presents \sys (\textbf{Neu}ro \textbf{S}ymbolic Link and \textbf{T}ime \textbf{I}nterval \textbf{P}rediction), the first comprehensive neuro-symbolic framework that addresses TKBC that encompasses both link prediction and time interval prediction tasks. We extensively evaluated the performance of \sys using benchmark time interval datasets and recorded a reasonable improvement in MRR metric as compared to that of the recent and notable baseline approaches. 
    
    \end{itemize}

\section{Related Work}

We classify the past literature on TKGC into four categories and discuss each of them in detail.

\subsection{Temporal KG Embedding Models}

A common thread of primitive TKGE approaches such as HyTE \cite{HYTE2018}, TA-TransE ~\cite{TA-TransE2018}, DA-TransE ~\cite{DA-TransE2020} is to encode the time information inside either the entity or relation embeddings and score a quadruple by employing a scoring function of a primitive model
 such as TransE~\cite{TransE2013} over the resulting time-aware embeddings.

However, recent models such as - TNTComplex~\cite{TNTComplex2020}, ChronoR ~\cite{ChronoR2021}, TeRo ~\cite{TeRO2020}, BoxTE~\cite{BoxTE2022} - have moved away from fundamental models as they adapt contemporary KGE models such as ComplEx ~\cite{complex2016}, RotatE ~\cite{Rotate2019}, BoxE~\cite{BoxE2020} to the temporal domain. 
In a different strand of research, models such as TransE-TAE~\cite{TransE-TAE2016}, Timeplex~\cite{Timeplex2020} - explicitly model the temporal constraints between a pair of tuples in TKG. 
Though effective, the above models learn black-box representations of embeddings that are not \textit{interpretable} to humans. Second, these approaches can not be \textit{generalized} to new data because the embeddings for the new data need to be relearned. Both these limitations are addressed by our proposed model.

\subsection{Temporal Multi-hop Reasoning Models} Temporal multi-hop reasoning models exploit neighborhood information of a query quadruple by employing distinct GNN architectures ~\cite{GNN2KipfandWelling017}. 
Models such as - TeMP~\cite{TeMP2020}, RE-NET ~\cite{RENET2020},  xERTE~\cite{XERTE2021}, CyGNet~\cite{CyGNet2020} exploit self-attention/GRU, RNN, time-aware attention mechanism, and Copy-Generation model respectively to integrate time information in a GNN.
Though interpretable to some extent \cite{XERTE2021}, these models are computationally expensive \cite{tace2022} and lack the generalizability to infer newer entities at the test time. 

\subsection{Temporal Rule Based Models}


Three recent temporal rule based models - TLogic~\cite{TLogic2021}, ALRE-IR~\cite{ALRE-IR2022}, TILP~\cite{TILP2023} - are closest to our work. TLogic and ALRE-IR models focus on link forecasting in the future and are designed for time instance datasets whereas our model can infer links in any time setting and is designed for time interval setting.
The closest to our work is - TILP~\cite{TILP2023} that performs all possible constrained walks on time interval datasets to learn temporal logic rules and adopts the attention mechanism to score each rule. Compared to TILP, the rule language of our model is entirely different and avoids linking unnecessary quadruples temporally in order to express the same information. 
Second, to the best of our knowledge, our proposed model is the first neuro-symbolic model that  performs time interval prediction for a given query.

\subsection{Time Prediction in TKGC}
Time prediction in TKG is relatively underexplored. Past research in this direction includes works such as Know-Evolve~\cite{KnowEvolve2017} and GHNN~\cite{GHNN2020} that perform \textit{time instance} prediction by modeling a given TKG fact as a point process. The closest to our work are - Time2Box~\cite{Time2Box2021} and Timeplex~\cite{Timeplex2020} models- that develop novel TKGE-based scoring functions to infer \textit{time interval} in TKG. Orthogonal to them, we exploit temporal rules to perform time interval prediction that endows our model with \textit{interpretability} while performing time interval prediction.

\section{Preliminaries}
\subsection{Temporal Knowledge Graphs}
\label{subsec:temporalknowledgegraphs}
A Temporal Knowledge Graph (TKG) is specified as $\mathcal{K} = \{\mathcal{E}, \mathcal{R}, \mathcal{Q}, \mathcal{T}\}$ where $\mathcal{E}$, $\mathcal{R}$ and $\mathcal{T}$  are the set of entities, relations and the domain of time instances respectively. A given TKG consists of a set of quadruples $\mathcal{Q} =\{(s_{n},r_{n},o_{n}, T_{n})\}_{n=1}^{\mid\mathcal{Q}\mid}$ where each quadruple ($s, r, o, T$) signifies that relation $r$ existing between subject entity $s$ and the object entity $o$ which is valid during the time interval $T$. Each time interval  $T \in \mathcal{T}\times \mathcal{T}$ is defined as $T$ = [$t_{b}, t_{e}$] 
 where $t_{b}$ and $t_{e}$ are the start and the end time of the time interval, respectively, and $t_{b} \leq t_{e}$. Additionally, $t_{min} \in \mathcal{T}$ and $t_{max} \in \mathcal{T}$ are the minimum and the maximum time instances attainable in a given TKG $\mathcal{K}$. To allow bi-directional walks in the model, it introduces an inverse link ($o, r^{-1}, s, T$) for every quadruple ($s, r, o, T$) present in the TKG.

\subsection{ Link and Time Interval prediction}
\label{subsec:linkandtimeintervalprediction}
The task of inferring missing links in a TKG is formulated as a query ($s, r, ?, T=[t_{b}, t_{e}]$). Consequently, in order to predict the subject entity in a quadruple, the query is expressed in terms of inverse relation as ($o, r^{-1}, ? , T= [t_{b}, t_{e}]$). 
Similar to the link prediction, the equally important task of time interval prediction is formulated as a query $(s,r,o,?)$ where the intent is to infer the time interval $T = [t_{b}, t_{e}]$ of the fact ($s,r,o$) in the TKG. 
\subsection{Allen Algebra }
\label{subsec:allenalgebra}
The presence of time intervals in the TKG necessitates the need for a formal mechanism that captures the temporal relations existing between the time intervals present in the data.
To that end, the model utilizes Allen's interval calculus~\cite{Allen1983} that encodes a total of $13$ possible relations between any two time intervals such that these relations are exhaustive and pairwise disjoint. For example, given two time intervals $T1=[t1_{b}, t1_{e}]$ and $T2=[t2_{b}, t2_{e}]$, Allen relation $\fol{overlaps(T1,T2)}$ holds between $T1$ and $T2$ if $t1_{b} < t2_{b} < t1_{e} < t2_{e}$. The proposed model utilizes all 13 Allen relations in its temporal rules. We refer to the 13 relations of Allen interval calculus as \textit{Allen relations} and the relation $\mathcal{R}$ introduced in Section \ref{subsec:temporalknowledgegraphs} as \textit{KG relations}.

\begin{figure*}[t]
\includegraphics[width=0.99\textwidth]
{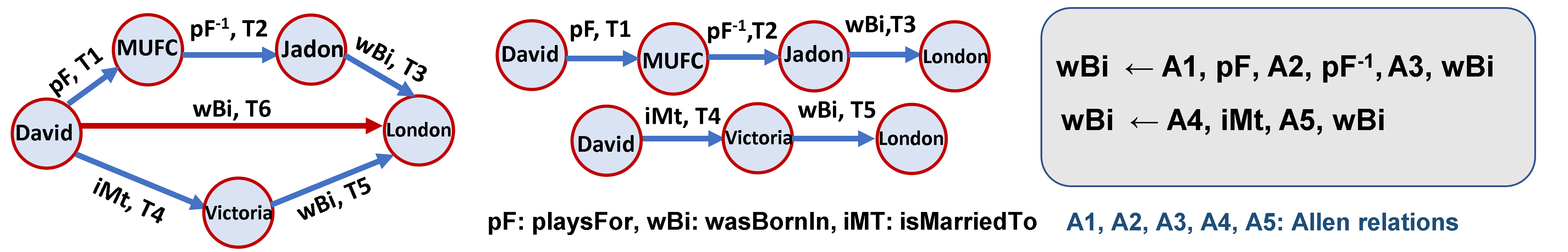}
\caption{Rule extraction on time interval data in the presence of Allen predicates. $\fol{wBi(David, London, T6)}$ is the target quadruple. The entity and time variables in the temporal rules on the right have been removed for brevity.}
\label{fig:AllWalks}
\end{figure*}

\section{Proposed Model}
\label{sec:proposedmodel}
In this section, we describe in detail the proposed temporal rules based framework, \textbf{\sys}, for link and time interval prediction in a given TKG. \textbf{\sys} first extracts \textit{all-walks} from TKG in the presence of Allen relations and cast them as temporal logic rules. It then learns the confidence score of a candidate answer for link and time interval prediction.
\subsection{Temporal Logic Rule Extraction}
\label{subsec:TemporalLogicRuleExtraction}
Given a quadruple ($s, r_{h}, o, T$) $\in \mathcal{K}$, \textbf{\sys} learns temporal rules of length $m$ such that ($s, r_{h}, o, T$) forms the rule head and rule body is captured by performing \textit{all-walks} on TKGs. In order to carry out \textit{all-walks}, it expresses TKG as graph $\mathcal{G}_{AW}$ wherein each quadruple is expressed as $\fol{s \xrightarrow{(r, \, T)} o}$ denoting an edge $\fol{(r,T)}$ between entities $\fol{s}$ and $\fol{o}$ (see Figure \ref{fig:AllWalks} (left)). It mines \textit{all-walks} over $\mathcal{G}_{AW}$ in three steps: $(a)$ First, beginning at $s$, it performs time-agnostic walks of length $m$ on $\mathcal{G}_{AW}$ such that the final node of the walk is $o$ (Figure \ref{fig:AllWalks} (middle)). It then expresses these walks in logical form. At this stage, these \textit{all-walks} exclusively consist of \textit{KG relations}. $(b)$ Next, it introduces \textit{Allen relations} into the \textit{all-walks} captured thus far, in order to bind the time intervals existing between neighboring \textit{KG relations} in the walk. Further, a special \textit{Allen relation} is introduced to bind the time interval of the target quadruple and the first $\textit{KG relation}$ in the walk. $(c)$ At the final step, it substitutes the constants with variables to generalize a grounded rule into a final rule (Figure \ref{fig:AllWalks}(right)). Formally, a temporal logic rule $L_{i} \in \mathcal{L}$ of length $m$ is defined as follows:
\begin{eqnarray}
r_{h}(E_{1}, E_{m+1}, T_{1}) &\longleftarrow \wedge_{i = 1} ^{m} \big(A_{i}(T_{i}, T_{i+1}) \nonumber \\ 
& \wedge r_{i}(E_{i}, E_{i+1}, T_{i+1})\big)
\label{eq:TemporalRuleDefinition}
\end{eqnarray}
where variable $E_{i} \in \mathcal{E}$, $T_{i} \in \mathcal{T}$, $A_{i}$ denote an entity, time interval, Allen relation and $r_{i} \in \mathcal{R}$ represents a fixed relation in TKG. Refer to Appendix \ref{appendix:ExampleoftemporalLogicRuleExtraction} for detailed examples.

Note that at step ($a$) of the model, when the model is currently at node $e_{i+1}$, it considers \textit{all} the edges originating from $e_{i+1}$ to its neighboring nodes and capture \textit{all} of them as resulting walks. This is because careful empirical analysis of TKG datasets reveals that the TKG interval datasets are sparser than TKG instance datasets and each rule generated from them can contribute to superior performance of the model. We call this mechanism of mining walks from TKG as '\textit{all-walks}'. 

\subsection{Score of the Candidate Answer}
\label{subsec:ScoreoftheCandidateAnswer}
Having learned the set of rules $\mathcal{L}$,  \textbf{\sys} now turns its attention towards estimating the confidence score of a candidate answer that it finds by grounding these rules. Recall that the goal of this paper is to accomplish both link prediction and time interval prediction.  Therefore, the candidate answer could be an entity during link prediction and two time instances: start time, $t_{b}$ and end time $t_{e}$ for a time interval prediction $T=[t_{b}, t_{e}]$. For instance, consider the link prediction task where the candidate entity $o$ can be arrived at by firing multiple rules in the rule set $\mathcal{L}$ and for one rule $L_{i} \in \mathcal{L}$, multiple \textit{paths} can be followed in the graph $\mathcal{G}_{AW}$ to arrive at entity $o$. Thus, for link prediction, its goal is to compute an overall confidence score, \textit{score$_{w}$(o)}, of candidate entity $o$ by the following formula:

\begin{align}
&score_{w}(o) = \sum_{L_{i} \in \mathcal{L}}score_{w}(o, L_{i})  \nonumber = \\
&  \sum_{L_{i} \in \mathcal{L}}\, \sum_{path \in \mathcal{P}(s, T,  L_{i}, o)}\psi_{w}(L_{i})[ob]\cdot\phi(path)[ob]
\label{eqn: finalscoreofcandidate}
\end{align}

where $\mathcal{P}(s, T,  L_{i}, o)$ is the set of grounded paths which start at ($s,T$) and end at $o$ following the rule $L_{i}$. Similarly, for time interval prediction, it learns two scores: $score_{w}(t_{b})$, the score of the starting time being $t_{b}$ and $score_{w}(t_{e})$, the score  of ending time being $t_{e}$. 
Note that each of the three scores has its own components for $\phi(path)$ and $\psi_{w}(L_{i})$. Without loss of generality, we explain the mechanism for computing the \textit{path score} $\phi(path)[ob]$ and \textit{rule score} $\psi_{w}(L_{i})[ob]$ for link prediction next (Section \ref{subsec:Pathscorecomputation} and \ref{subsection: ruleweightcomputation}), while only explaining the mechanism for time-prediction where it fundamentally differs from link prediction.

\subsection{Path Score Computation}
\label{subsec:Pathscorecomputation}
\paragraph{Link Prediction:} Although multiple heuristics can be explored to score the path, $\phi(path)[ob]$, for link prediction, the model exploits a relatively simple approach and sets $\phi(path)[ob] = 1$ for each $path$ that reaches the target answer $o$ when instantiating a given rule $L_{i}$.

\paragraph{Time Interval Prediction:} In order to develop the heuristic  for \textit{path score}  for a candidate start time instance $t_{b}$ - $\phi(path)[t_{b}]$, reconsider a rule $L_{i}$ of length $2$ in Equation \ref{eq:TemporalRuleDefinition}. 
While we ground the rule $L_{i}$ for query ($s,r,o, ?$), we observe that the rule is partially grounded because $T_{1}$ remains unknown. This entails two key challenges: first, the potential set of candidate start time instances, $t_{b}$, remain unknown. Second, $\phi(path)[t_{b}]$ needs to be determined by leveraging only the partially-grounded path.  
To overcome these challenges,  the model considers the first Allen predicate $A_{1}$ and the grounding - $r_{1}(s, x, T_{2}=[t_{2b}, t_{2e}])$- of the first \textit{KG relation} $r_{1}$ in the partially grounded path.  It exploits $A_{1}$ and $r_{1}$ in identifying the set of potential candidates - [$t_{1s}, t_{2s}$] - of the start time $t_{b}$. For example, if $A_{1}$ is Allen predicate $\fol{before}$, then the set [$t_{1s}, t_{2s}$] for start time instance is  [$t_{min}, t_{2b}$).

Next, to evaluate $\phi(path)[t_{b}]$ for a given $t_{b} \in [t_{1s}, t_{2s}]$, it considers a relation pair ($r_{h}, r_{1}$) and maintains parameters $\mu_{r_{h}r_{1}}^{start}$ and $\sigma_{r_{h}r_{1}}^{start}$ whose purpose is to store the mean and standard deviation of the difference of start time of relation $r_{h}$ and $r_{1}$. Then, it obtains a score $\phi(t_{b})$  by modeling the time gap between the start time at the rule head ($t_{b}$) and the start time of the first \textit{KG relation} $r_{1}$ in the rule body ($t_{2b}$) as Gaussian distribution, as follows:
\begin{align}
\phi(t_{b}) = \mathcal{N}(t_{b}- t_{2b}\vert \mu_{r_{h}r_{1}}^{start}, \, \sigma_{r_{h}r_{1}}^{start})
\end{align}
 
The final \textit{path score} value of $t_{b}$, $\phi(path)[t_{b}]$, is computed by normalizing the score $\phi(t_{b})$ with respect to all the potential start time instances in [$t_{1s}, t_{2s}$].
Likewise, it computes $\phi(path)[t_{e}]$ for the end time instance $t_{e}$ by considering the statistics of end time $t_{2e}$ of first \textit{KG relation} $r_{1}$ in rule body.
\subsection{Rule Score Computation}
\label{subsection: ruleweightcomputation}
We now delve into the estimation of the \textit{rule score}, $\psi_{w}(L_{i})$, which is the second key component of $score_{w}(o)$ (Equation \ref{eqn: finalscoreofcandidate}). 
It is worth noting that the model learns an exclusive \textit{rule score} for each rule that does not rely on rule groundings (or \textit{paths}). Therefore, it disregards the entities and time variables in a given rule $L_{i}$ and perceives it as $L_{i}: \, r_{h} \leftarrow \wedge_{t=1}^{m} A_{t} \wedge r_{t}$ that forms the basis of $\psi_{w}(L_{i})$'s estimation. The \textit{rule score} estimation has three main ingredients which are discussed below.
\subsubsection{Logical Path Embeddings}
\label{subsub: LogicalPathEmbeddings}
The model first learns a unique embedding representation for the body $\wedge_{t=1}^{m} A_{t} \wedge r_{t}$ of a given rule $L_{i}$. It is called \textit{path embedding} and denoted as $\textbf{p}_{L_{i}}$. Motivated by ARLE-IR model~\cite{ALRE-IR2022}, we employ Gated Recurrent Unit (GRU) model~\cite{GRU2014} for path embedding learning. At time $t$, the input of the form $\textbf{x}_{t} = [\textbf{A}_{t}; \, \textbf{r}_{t}]$ is fed to GRU where $\textbf{A}_{t}$ and $\textbf{r}_{t}$ are the embedding vectors of t-th \textit{Allen predicate} $A_{t}$ and \textit{KG relation} $r_{t}$ in a rule $L_{i}$'s body. A GRU unit utilizes the following function in order to generate the hidden-layer embedding $\textbf{h}_{t}$ at time $t$:
\begin{align}
\textbf{r}_{t} &= \sigma\big(\textbf{W}_{r}\cdot\textbf{x}_{t} \, + \, \textbf{U}_{r}\cdot\textbf{h}_{t-1}\, + \, \textbf{b}_{r} \big) \\
\textbf{z}_{t} &= \sigma\big(\textbf{W}_{z}\cdot\textbf{x}_{t} \, + \, \textbf{U}_{z}\cdot\textbf{h}_{t-1} \, + \, \textbf{b}_{z} \big) \\
\textbf{n}_{t} &= \tanh \big(\textbf{W}_{n}\cdot \textbf{x}_{t} \, + \, \textbf{r} \odot \textbf{h}_{t-1} \, + \, \textbf{b}_{n}\big) \\
\textbf{h}_{t} &= (1-\textbf{z}_{t})\odot \textbf{n}_{t} \, + \, \textbf{z}_{t} \odot \textbf{h}_{t-1}
\end{align}
where $\textbf{r}_{t}$ is the \textit{reset} gate that allows the hidden state to discard information that is insignificant in the future and $\textbf{z}_{t}$ is the \textit{update} gate that controls how much information from $\textbf{h}_{t-1}$ is carried over to $\textbf{h}_{t}$. The final hidden state embedding $\textbf{h}_{m}$ after $m$ sequential steps of GRU represents the path embedding $\textbf{p}_{L_{i}}$ of a given rule $L_{i}$.

\subsubsection{Similarity Matching Evaluation}
\label{subsubsection:RuleConfidenceEvaluation}
After retrieving the path embedding $\textbf{p}_{L_{i}}$ of a given rule $L_{i}$, the model next focuses on evaluating the \textit{similarity score} by accessing the similarity between the path and the rule head embeddings. Because the objective of the model is to accomplish both link prediction and time interval prediction tasks, it considers three vectors - $\textbf{r}^{h_{i}}_{o}$, $\textbf{r}^{h_{i}}_{t_{b}}$ and $\textbf{r}^{h_{i}}_{t_{e}}$ - for relation head $r_{h}$ that signify the link prediction task and start time, $t_{b}$, and end time, $t_{e}$, of a given time interval prediction task respectively. Then, it estimates three \textit{similarity scores} by encoding the interaction between the path embedding and rule head embedding components as cosine similarity function. For example, the \textit{similarity score} between $\textbf{p}_{L_{i}}$ and $\textbf{r}^{h_{i}}_{o}$ for link prediction is defined as:
\begin{equation}
  f(\textbf{p}_{L_{i}}, \textbf{r}^{h_{i}}_{o}) = cos(\textbf{p}_{L_{i}}, \textbf{r}^{h_{i}}_{o})  
\end{equation}
Two other scores $f(\textbf{p}_{L_{i}}, \textbf{r}^{h_{i}}_{t_{b}})$ and $f(\textbf{p}_{L_{i}}, \textbf{r}^{h_{i}}_{t_{e}})$ are computed analogously. Finally, we normalize this score so that it lies in the range [0,1].

\subsubsection{PCA Score of Rule}
\label{subsubsec:PCAScoresComputation}
In addition to  \textit{similarity score} $f(\textbf{p}_{L_{i}}, \textbf{r}^{h_{i}}_{\cdot})$, the model leverages PCA score~\cite{AMIE2013} for each rule $L_{i}$ denoted as $PCA(L_{i})$. This symbolic rule confidence metric acts as a prior of the rules and helps in informed initialization of the \textit{rule score} $\psi_{w}(L_{i})$.  It is to be noted that $PCA(L_{i})$ is not a learnable parameter and is evaluated once for each rule at the beginning of the training. (See Appendix \ref{appendix:PCAScore} for PCA score). The final \textit{rule score} $\psi_{w}(L_{i})[ob]$ Equation \ref{eqn: finalscoreofcandidate} is defined as:
\begin{equation}
   \psi_{w}(L_{i})[ob] =    f(\textbf{p}_{L_{i}}, \textbf{r}^{h_{i}}_{o}) * PCA(L_{i})[ob]
\end{equation}
The \textit{rule scores} - $\psi_{w}(L_{i})[t_{b}]$ and $\psi_{w}(L_{i})[t_{e}]$ - of start and end time instances, $t_{b}$ and $t_{e}$ can be estimated in a similar manner by replacing the similarity score $f()$ and PCA score $PCA(L_{i})[\cdot]$ in above equation with analogous values. 

\subsection{Loss Function}
\label{subsec: lossfunction}
For a given quadruple ($s,r,o,T$=$[t_{b}, t_{e}]$) $\in \mathcal{K}$, the model generates two queries: ($s,r,?,T$) for the link prediction and ($s,r,o, ?$) for the time interval prediction. It utilizes two loss functions to optimize the model for the two sub-goals: $\mathcal{L}_{LP}$ for link prediction and $\mathcal{L}_{TP}$ for the time interval prediction. We explain each of them in detail next.

\subsubsection{Loss Function for Link Prediction, $\mathcal{L}_{LP}$}
\label{subsubsec: lossfunctionforlinkprediction}
The proposed model first computes the probability $P(s,r,o,T)$ of arriving at a candidate answer $o$ for a given query ($s,r,?, T$) by employing the softmax function on $score_{w}(o)$ over all the entities $\mathcal{E}$.
We use the notation $P(o)$ to denote $P(s,r,o,T)$ for link prediction. For a given query, let $\mathcal{D}$ represents set of \textit{true} objects  such that $\forall o \in \mathcal{D}$, the quadruple $(s,r,o,T) \in \mathcal{K}$ and the set $\mathcal{N}$ = $\mathcal{E}\setminus\mathcal{D}$ represents the \textit{false} objects such that $\forall f \in \mathcal{N}$,  the quadruple $(s,r,f,T) \not\in \mathcal{K}$. Also, let $S(o)$ represents a set of \textit{false} objects that generate higher scores than a \textit{true} object $o$ where $S(o) \subset N$. Then, the loss function for link prediction, $\mathcal{L}_{LP}$, is defined as:

\begin{equation}
 \sum_{f \in \mathcal{N}}P(f) + \sum_{o \in \mathcal{D}}\Big(\dfrac{\sum_{n \in S(o)}(P(n)-P(o))}{\vert S(o)\vert}\Big)
\end{equation}
\subsubsection{Loss Function for Time Interval, $\mathcal{L}_{TP}$}
For time interval prediction, the model considers two scores: $score_{w}(t_{b})$ and $score_{w}(t_{e})$ and employs softmax on them over all the time instances $\mathcal{T}$ retrieved from the TKG to obtain $P(s,r,o,t_{b})$ and $P(s,r,o,t_{e})$ respectively. For short, we denote these probabilities as $P(t_{b})$ and $P(t_{e})$ for time prediction. For a given query ($s,r,o, ?$), let $\mathcal{D}$ represent the set of \textit{true} intervals such that for $T \in \mathcal{D}$, $(s,r,o,T) \in \mathcal{K}$. We now divide $\mathcal{D}$ into two sets: $\mathcal{D}_{b}$ and $\mathcal{D}_{e}$, that represent the start and the end time instance in $\mathcal{D}$. Now, $\mathcal{N}_{b}$ = $\mathcal{T}\setminus \mathcal{D}_{b}$ and $\mathcal{N}_{e}$ =$\mathcal{T}\setminus \mathcal{D}_{e}$ represent \textit{false} start time and \textit{false} end time for a query respectively. Because time instances $t_{b}$ and $t_{e}$ are numerical in nature, the model endows the loss function with numeric difference capability and defines the final loss $\mathcal{L}_{TP}$ as follows:
\begin{align}
&\sum_{t_{b} \in \mathcal{D}_{b}}\sum_{f_{b} \in \mathcal{N}_{b}}\big(P(f_{b})- P(t_{b})\big)\ast d\big( f_{b} - t_{b}\big) + \nonumber \\
& \sum_{t_{e} \in \mathcal{D}_{e}}\sum_{f_{e} \in \mathcal{N}_{e}}\big(P(f_{e})- P(t_{e})\big)\ast d\big( f_{e} - t_{e}\big)
\end{align}
where distance $d$ between \textit{true} and the \textit{false} time instances and is explained in Appendix \ref{appendix: distancebetweentimeinstances}. 
\subsection{Extended Model with Timeplex}
\label{subsec:extendedmodelwithtimeplex}
Though appealing, logic-based models are generally limited in modeling the numeric features present in the KG~\cite{NeuralNumLP2020}. To overcome this limitation in our work, we extend our model by explicitly capturing two numeric features: ($i$) relation recurrence feature ($ii$) relation pairs feature, similar to Timeplex~\cite{Timeplex2020}. \textit{Relation recurrence} feature captures the distribution of time difference between the recurrences of a given relation whereas \textit{relation pairs feature} captures the distribution of the time gap between the start time of pair of relations. Time distributions in both features are modeled as Gaussian distribution.  

Additionally, we experiment with ensembling our model with \textit{Timeplex (base)}~\cite{Timeplex2020} in order to observe the effect of integrating the complementary features of temporal rule-based models with time-aware embedding-based model. 
\section{Experiments}
\label{sec: experiments}
We investigate the following research questions in our experiments:
(1) Does our model outperform  both the best embedding based and the best rule-based systems for the task of link prediction?
(2) Does our model outperform the best embedding-based models for the task of time interval prediction?
(3) Do the rules with temporal constraints in the form of \textit{Allen relations} help in improving the link prediction performance? (4) Are the rule generated by our proposed model \textit{human-interpretable}?

\subsection{Datasets and Experimental Setup}
\textbf{Datasets: } We evaluate the proposed model on two standard TKBC datasets. WIKIDATA12k and YAGO11k ~\cite{HYTE2018} are two temporal knowledge graphs that have a time interval associated with each triple $(s,r,o)$. An example of facts contained in these datasets is ($Alice\; Hamilton, graduatedFrom, University$ $  of \;Chicago, [1899,1901]$). For both these datasets, we consider the temporal granularity to be $1$ year. Since month and day are not present in the majority of the samples, we discard month and day information from these datasets that results in a more uniform representation. Statistics for these datasets are presented in Table \ref{tab:KG statistics} in Appendix \ref{appendix: datastatistics}.
\paragraph{Algorithms compared:} We compare the performance of our model on link prediction to the static and temporal rule-based models and static and temporal embedding-based models. The baselines for static rule-based models comprise of Neural-LP~\cite{NeuralLP2017}, and AnyBurl~\cite{AnyBURL2019}. Further, TLogic~\cite{TLogic2021} and TILP~\cite{TILP2023} serve as a baseline for the temporal rule-based model. In addition, we compare against ComplEx~\cite{complex2016} which is a static embedding-based model. Besides, we consider five temporal embedding-based models which are TA-ComplEx~\cite{TA-TransE2018}, HyTE~\cite{HYTE2018}, DE-SimplE~\cite{DA-TransE2020}, TNT-Complex~\cite{TNTComplex2020}, and TimePlex~\cite{Timeplex2020}. 

For time interval prediction, we compare the performance of the proposed model against the temporal embedding-based models which are HyTE~\cite{HYTE2018}, TNT-Complex~\cite{TNTComplex2020}, and TimePlex~\cite{Timeplex2020}. For all the time interval prediction models and link prediction of embedding-based models, results are directly taken from Jain et al. (\citeyear{Timeplex2020}). Similarly, for the rule-based models for link prediction, the baseline results are taken from Xiong et al. (\citeyear{TILP2023}).

We compare three variants of our proposed model with the baselines. First, we compare our model which is trained exclusively with the proposed temporal rules and corresponding candidate score in Equation \ref{eqn: finalscoreofcandidate} and name this variant as our base model and denote it as `\sys (Base)'. Next, we integrate the \textit{relation recurrence} and \textit{relation pairs} numerical features (Section \ref{subsec:extendedmodelwithtimeplex}) into the score of the base model and name this variant as `\sys with Gadgets'. Finally, in the third variant, we add the score of both the gadgets and Timeplex (Base) into the score of our base model and call this variant as `\sys with KGE'.   

\paragraph{Experimental Details:} For all the results reported for the proposed model, we optimize parameters of the loss in Section \ref{subsec: lossfunction} with an Adam optimizer \cite{Adam2015} while decreasing the learning rate by Cosine Annealing ensuring that the minimum learning rate at any time during the training does not fall below the one-fifth of its initial value. To get the best results for link prediction, we train our model for $5000$ epochs. Likewise, for time interval prediction,  we train the model for $2000$ epochs. We set a dimensionality for all the \textit{Allen relation} embeddings, \textit{KG relation} embeddings, and the rule head embeddings (which have the same dimension as the hidden dimension of the GRU) to be $32$. Besides, we set the maximum rule length as 3 for both datasets.

We utilize the standard metrics of Mean Reciprocal Rank (MRR), Hits@1, and Hits@10 for comparison in the link prediction task. Similar to Jain et al.(~\citeyear{Timeplex2020}), we evaluate our model with time-aware filtering because it results in more valid performance estimation. For time interval prediction, we employ the aeIOU metric for comparison 
 that was introduced by Jain et al. (\citeyear{Timeplex2020}). During the training of the model, we select the best validation model for link prediction based on the MRR metric and the best validation model for time interval prediction based on the aeIOU metric. Further details of the hyperparameters adopted for all the experiments is provided in Appendix \ref{appendix:hyperparametersoftheexperiments}.

\begin{table}[t]
 \caption{Results of link prediction on two datasets. Baseline results taken from TILP paper }
\label{tab:tablemainlinkprediction}
\centering
\small
\begin{center}

\resizebox{\columnwidth}{!}{
\begin{tabular}{|p{2.6cm}|p{0.60cm}p{0.55cm}p{0.70cm}|
p{0.60cm}p{0.55cm}p{0.70cm}|}
\Xhline{3\arrayrulewidth}
\multirow{2}{2em}{\textbf{Algorithm}} & \multicolumn{3}{c|}{\textbf{WIKIDATA12k}} &
\multicolumn{3}{c|}{\textbf{YAGO11k}}\\
& MRR & H@1 & H@10 & MRR & H@1 &  H@10  \\
\Xhline{3\arrayrulewidth}
Neural-LP & 18.23 &  9.08 & 38.48 & 10.01 & 4.01 & 18.45 \\
AnyBURL & 19.08 & 10.30 & 39.04 & 9.08 & 3.78 & 18.14 \\
TLogic & 25.36 & 17.54 & 44.24 & 15.45 & 11.80 & 23.09 \\
TILP-base & 31.14 & 21.52 & 50.77 & 18.80 & 13.36 & 30.89 \\
TILP & 33.28 &23.42 & 52.89 & 24.11 & 16.67 & \textbf{41.49} \\
\Xhline{3\arrayrulewidth}
ComplEx & 24.82 & 14.30 & 48.90 & 18.14 & 11.46 & 31.11 \\
TA-ComplEx & 22.78  & 12.69  & 46.00 & 15.24  & 9.36 & 26.26 \\
HyTE  & 25.28   & 14.70  &  48.26 & 13.55  & 3.32   &  29.81\\
DE-SimplE & 25.29 & 14.68 & 49.05 & 15.12 & 8.75 & 26.74   \\
TNT-Complex & 30.10 & 19.73 & 50.69 & 18.01 & 11.02 & 31.28 \\
TimePlex (Base) & 32.38 & 22.03 & 52.79 & 18.35 & 10.99 & 31.86 \\
TimePlex & 33.35 & 22.78 & 53.20 & 23.64 & 16.92 & 36.71 \\
\Xhline{3\arrayrulewidth}
\sys (Base)& 31.17 & 21.03 & 50.15 & 23.84 & 17.36 & 34.86  \\
\sys w/ Gadgets & \textbf{34.27}& \textbf{23.99} & 53.05 & 25.21& 18.04 & 38.05 \\
\sys w/ KGE & 34.15 & 23.97& \textbf{53.17} & \textbf{25.36} & \textbf{18.16} & 38.27 \\
\Xhline{3\arrayrulewidth}
\end{tabular}}
\end{center}
\end{table}

\subsection{Results and Observations}
\paragraph{Link Prediction}
Table \ref{tab:tablemainlinkprediction} compares our proposed model against all the base algorithms for the task of link prediction. We observe that the performance of our base model (performance of rules only) is comparable with the base model of the best embedding-based model (TimePlex (Base)) for WIKIDATA12k, while the gap is more pronounced on YAGO11k as our base model outperforms TimePlex (base) on all the three metrics (gain over 5 MRR pts). Similarly, our base model is comparable to the base model of the best temporal rule-based model (TILP-base) for WIKIDATA12k whereas our base model outperforms it on all three metrics on YAGO11k dataset. We also observe that our model, when integrated with gadgets (\sys w/ Gadgets), outperforms the state-of-the-art models on all the metrics on YAGO11k, and 2 out of 3 metrics on WIKIDATA12k. Our model shows a similar trend in performance when it is ensembled with the complete TimePlex model (\sys w/ KGE).

\begin{table}[H]
 \caption{Results of time interval prediction on two datasets. The baseline results are taken from Timeplex}
\label{tab:tabletimeprediction}
\centering
\small
\begin{center}
\begin{tabular}{|l|P{1.95cm}|P{1.5cm}|}
\Xhline{3\arrayrulewidth}
\textbf{Dataset} & \textbf{WIKIDATA12k} & \textbf{YAGO11k} \\
\Xhline{3\arrayrulewidth}
Algorithm & aeIOU & aeIOU \\
\Xhline{3\arrayrulewidth}
HyTE & 5.41 & 5.41 \\
TNT-Complex & 23.35& 8.40 \\
Timeplex (Base) & 26.20 & 14.21 \\
Timeplex  & 26.26 & 20.03 \\
\Xhline{3\arrayrulewidth}
\sys (Base) & 26.27  & 16.42 \\
\sys w/ Gadgets &  26.30 & \textbf{26.35}  \\
\sys w/ KGE & \textbf{27.35} &  24.88 \\
\Xhline{3\arrayrulewidth}
\end{tabular}
\end{center}
\end{table}

\paragraph{Time Interval Prediction}
Table \ref{tab:tabletimeprediction} compares our model with all base models for the task of time interval prediction. To the best of our knowledge, we are the first to investigate the task of predicting time intervals in neuro-symbolic TKGC setting. We observe that the performance of our base model (\sys (Base)) is comparable with the base model of the state-of-the-art model (TimePlex (Base)) for WIKIDATA12k, while our model performs around 2 points (aeIOU) better on the YAGO11k dataset. Further, when integrated with gadgets, our model shows superior results by outperforming the state-of-the-art model (TimePlex) on YAGO11k by a significant margin (around 6 aeIOU pts), while still giving  a decent performance on the WIKIDATA12k dataset. Finally, our model showcases the best performance for WIKIDATA12k when it is ensembled with TimePlex. Note that the trend is similar when gadgets are employed in TimePlex(Base) and \sys (base)  for the WIKIDATA12k dataset because gadgets do not assist any of the two models in yielding better performance.



\subsection{Diagnostics}
\paragraph{Human interpretability of logical rules}
One advantage of our temporal rule-based model is that the predictions are in human-interpretable form, whereas the predictions of embedding-based models are opaque in nature. Here, we demonstrate some real examples of how the rules in  \sys model help in predicting the correct entity/time interval in Yago11k dataset:

\noindent \textbf{Query:} (Franz Dahlem, isAffiliatedTo, ?, [1920,1946])

\noindent \textbf{Correct Answer:} Communist Party of Germany 

\noindent \textbf{The rule that grounds the gold object:} 

\noindent $isAffiliatedTo(E1,E2,T1) \leftarrow During (T1,$ $ T2) \wedge isMarriedTo(E1,E3,T2) \wedge Contains$ $ (T2,T3)\land isAffiliatedTo(E3,E2,T3)$

\noindent \textbf{The groundings:} $E1:$ Franz Dahlem, $E2:$ Communist Party of Germany, $E3:$ Kathe Dahlem, $T1:$ [1920, 1946], $T2:$ [1899, 1974], $T3:$ [1920, 1946]\\

\noindent The above rule provides an \textit{explanation} of why \textit{Franz Dahlem} was affiliated to \textit{Communist Party of Germany} at a given time interval by reasoning that his wife \textit{Kathe Dahlem} was also affiliated to this party during their marriage. We provide another example of interpretable rule from YAGO11k dataset for time prediction as below.\\

\noindent \textbf{Query:} (Donna Hanover, isMarriedTo, Rudy Giuliani, ?)\\
\textbf{Correct Answer:} [1984,2002]\\
\textbf{Rule grounding gold start and gold end:}\\
$isMarriedTo(E1,E2,T1) \leftarrow Equals (T1, T2) $ $  \wedge isMarriedTo^{-1}(E1,E2,T2)$ \\
\textbf{The Grounding:} $E1:$ Donna Hanover, $E2:$ Rudy Giuliani, $T2:$ [1984, 2002]\\

The above rule provides us the \textit{temporal information} that since \textit{Rudy} was married to \textit{Donna} for a given time interval, \textit{Donna} was also married to \textit{Rudy} for the exact same time interval due to the symmetric nature of the relation. Similar examples for WIKIDATA12k are presented in Appendix \ref{appendix:humaninterpretabilityofrules}.
\begin{table}[H]
 \caption{Link prediction performance with/without temporal constraints in rules}
\label{tab:tabletemporalconstraints}
\centering
\small
\begin{center}
\resizebox{\columnwidth}{!}{
\begin{tabular}{|l|p{0.55cm}p{0.55cm}p{0.70cm}|
p{0.55cm}p{0.55cm}p{0.70cm}|}
\Xhline{3\arrayrulewidth}
\multirow{2}{2em}{\textbf{Algorithm}} & \multicolumn{3}{c|}{\textbf{WIKIDATA12k}} &
\multicolumn{3}{c|}{\textbf{YAGO11k}}\\
& MRR & H@1 & H@10 & MRR & H@1 &  H@10  \\
\Xhline{3\arrayrulewidth}
\sys & \textbf{31.17} & \textbf{21.03} & \textbf{50.15} & \textbf{23.84} & \textbf{17.36} & \textbf{34.86}  \\
\sys - TR & 26.00 & 15.60 & 47.14 & 18.90& 11.68 & 30.40 \\
\Xhline{3\arrayrulewidth}
\end{tabular}}
\end{center}
\end{table}

\paragraph{Advantage of Temporal Constraints in Rules}
Next, we investigate the importance of incorporating Allen relations into the temporal rules learned by our proposed model. Table \ref{tab:tabletemporalconstraints} compares the performance our base model for link prediction with the performance of a variant of our model where we employ the rules by discarding the Allen relations in them. For both cases, we use the same set of hyperparameters as reported in Table \ref{tab: hyperparameterslinkprediction} in Appendix \ref{appendix:hyperparametersoftheexperiments} (base model). In order to train the model for rules without the Allen constraints, we can essentially treat the absence of Allen relation as a `NOREL' constraint, and we feed the embedding of this at every step in the input $x_{t}$ to the GRU instead of the Allen relation embeddings. We name this model as `\sys-TR' and present the results in Table \ref{tab:tabletemporalconstraints}. We observe that the presence of Allen relation constraints is critical for our model because it helps the model in improving its performance substantially on both the datasets.

\section{Conclusion}
In this paper, we develop a novel neuro-symbolic TKGC method that represents the temporal information of TKGs as a unique
temporal rule language and ascribe a confidence score to a candidate answer, which can be an entity or time interval,
by leveraging a newly designed scoring function. The key strength of the proposed formulation is that it can perform both link prediction and time interval prediction in neuro-symbolic setting. Compared to previous methods, our model has made substantial improvement in both link prediction and time interval prediction over two benchmark datasets. Furthermore, we show that our model results in human-interpretable reasoning while answering a given link prediction and time prediction query.

\section*{Limitations}

One limitation of rule based models, in general, is that they cannot directly capture the numeric features present in the data and models generally rely on some supplementary mechanism to harness the numeric features. Consequently, we exploited the gadgets proposed in Timeplex~\cite{Timeplex2020} to capture numeric features present in TKGs.



\section*{Ethics Statement}
 \label{sec:ethicsstatement}
 We anticipate no substantial ethical issues arising due to our work on link prediction and time interval prediction for Neuro-Symbolic TKGC.


\bibliography{anthology,custom}
\bibliographystyle{acl_natbib}

\appendix

\begin{figure*}[t]
\centering
\includegraphics[scale = 0.5]
{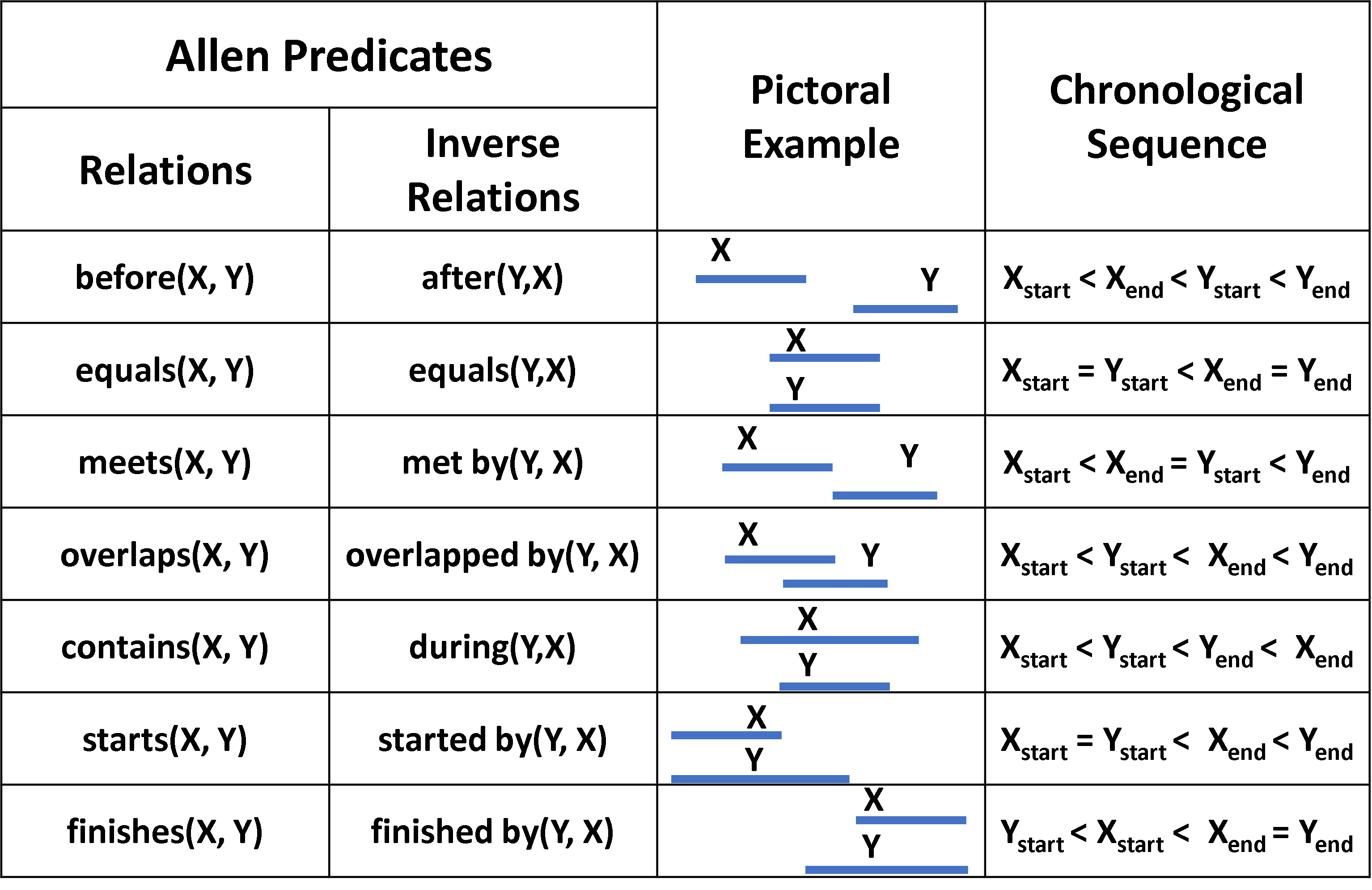}
\caption{This figure lists all the 13 relations in Allen algebra calculus ~\cite{Allen1983}. The pictorial example in the third column is for the relations in the first column}
\label{fig:AllenPredicates}
\end{figure*}


\section{Allen's Interval Calculus}
The quadruples considered in our setting encode time intervals in the last argument, requiring the use of Allen relations as a formal technique that captures the temporal relations between the time intervals present in the data.
As discussed in Section \ref{subsec:allenalgebra}, we utilize all the $13$ relations possible between any two time intervals in our temporal rules. In this section, we describe all 13 Allen relations in detail in Figure \ref{fig:AllenPredicates}. Each of the relations in Allen algebra calculus can be written as set of rules. For example If we have one time interval $X = [X_{start}, X_{end}]$ and another time interval $Y = [Y_{start}, Y_{end}]$ then the Allen relation before exists between them, i.e. $\fol{before(X,Y)}$ iff $X_{start} < X_{end} < Y_{start} < Y_{end}$. Similarly, constraints for all the relations is defined in the last column named `Chronological Sequence' of Figure \ref{fig:AllenPredicates}

\section{An Example of Temporal Logic Extraction}
\label{appendix:ExampleoftemporalLogicRuleExtraction}
We consider an example of temporal logic rule extraction based upon three steps defined in Section \ref{subsec:TemporalLogicRuleExtraction}. Our example is based upon a fragment of Yago11k dataset which is shown in Figure \ref{fig:AllWalks}. In this example $\fol{pF}, \fol{pF^{-1}}, \fol{wBi}, \fol{iMT}$ denote shortcuts for the relations $\fol{playsFor}$, $\fol{playFor^{-1}}$, $\fol{wasBornIn}$, $\fol{isMarriedTo}$ respectively. The relations $\fol{A1}$ to $\fol{A5}$ represent Allen relations. In order to learn a rule that is based upon target quadruple $\fol{wBi(David, London, T6)}$ the model would first obtain a walk $\fol{David}$ $ \fol{\xrightarrow{(iMt, T4)} Victoria \xrightarrow{(wBi, T5)} London}$ on $\mathcal{G}_{AW}$. Further, this walk would be expressed in the logical form as  $\fol{iMt(David, Victoria,T4) \wedge}$ $\fol{ wBi(Victoria, London,T5)}$. 

In the next step (step (b) in Section \ref{subsec:TemporalLogicRuleExtraction}), Allen predicates are introduced into the walk,  the corresponding example would be expressed as $\fol{A4(T6,T4) \wedge iMt(David, Victoria, T4) \wedge A5(T4}$ $\fol{,T5) \wedge wBi(Victoria, London, T5)}$. Please note that these Allen relations denote one of the 13 relations in Allen Algebra calculus.  The final rule after introducing variables is $\fol{wBi(A,B,C) \leftarrow}$ $\fol{  A4(C,F) \wedge iMt(A,D,F) \wedge A5(F,G) \wedge wBi(D,B,G)}$. This rule is expressed without the entity and time interval variables in Figure \ref{fig:AllWalks}(right) as $\fol{wBi \leftarrow}$ $\fol{A4 \wedge iMt \wedge A5 \wedge wBi}$

\section{PCA Score metric for temporal data}
\label{appendix:PCAScore}

PCA metric\cite{AMIE2013} is based on the Partial Closed World assumption according to which if we know one object $o$ for a given $s$ and $T$ in a quadruple ($s,r_{h},o, T$) then we know all the $o'$ for that $s$ and $T$. If we consider temporal rule $L_{i}$ to be $\fol{B} \Rightarrow \fol{r_{h}}(\fol{s}, \fol{o}, \fol{T})$, the PCA score of this rule for link prediction, $PCA(L_{i})[ob]$, is:

\scalebox{0.87}{\parbox{.5\linewidth}{%
\begin{align*}
    \label{eq: PCAScore}  
     \fol{\dfrac{\#(s,o, T) :  |N(s, B, o, T)| > 0 \wedge r_{h}(s,o, T) \in P}{\#(s,o,T) : |N(s, B, o,T)| > 0 \wedge \exists o': r_{h}(s,o',T) \in P}}
\end{align*}
}}

Here, $\fol{N(s,B,o,T)}$ denotes the path in the body $\fol{B}$ of the rule $L_{i}$. This implies that we divide the number of positive examples $\fol{P}$ satisfied by the rule  by the total number of ($s,o,T$) satisfied by the rule such that $r_{h}(s,o',T)$ is a positive example for some $o'$. Similarly, we define the PCA score for start time instance $t_{b}$, $PCA(L_{i})[t_{b}]$, as

\scalebox{0.85}{\parbox{.5\linewidth}{%
\begin{align*}
     \fol{\dfrac{\#(s,o, t_{b}) :  |N_{tb}| > 0 \wedge \exists t_{e} \in T, r_{h}(s,o, [t_{b}, t_{e}]) \in P}{\#(s,o,t_{b}) : |N_{tb}| > 0 \wedge \exists T': r_{h}(s,o,T') \in P}}
\end{align*}
}}

$|N_{tb}|$ is a notation for $|N(s, B, o, t_{b})|$. This implies that we divide the number of positive examples $\fol{P}$ satisfied by the rule  by the total number of ($s,o,T$) satisfied by the rule such that $r_{h}(s,o,T')$ is a positive example for some $T'$.
\section{Distance computation between time instances}
\label{appendix: distancebetweentimeinstances}

In order to find the distance $\fol{d}$  between two time instances $\fol{t_{a}}$ and $\fol{t_{b}}$ i.e. $\fol{d\big( t_{a} - t_{b}\big)}$, the model sorts the years in $\mathcal{T}$ in increasing order and assign a unique id to each of them. The difference $\fol{d}$ is then taken between those ids'. The difference is then divided by the maximum difference between any two ids, in order to follow the constraint that $0 < d(.) < 1$.

\section{Data Statistics}
\label{appendix: datastatistics}
The details of the datasets used for experimentation in Section \ref{sec: experiments} are provided in Table \ref{tab:KG statistics}. We utilize two datasets - YAGO11k and WIKIDATA12k for our experimentation. Both these datasets are time interval-based datasets.

\begin{table}[H] 
\centering
\small
\caption{Statistics of Temporal Knowledge Graph datasets}
\vspace{0.03in}
\begin{tabular}{|c|c|c|}
\Xhline{3\arrayrulewidth}
Features & YAGO11k & WIKIDATA12k \\
\hline
\#Entities &10622 & 12554 \\
\#Relations &10 & 24 \\
\#Instants & 251 &237\\
\#Intervals & 6651 &2564\\
\#Training &16408 & 32497\\
\#Validation &2051 & 4062\\ 
\#Test  & 2050  & 4062 \\
\Xhline{3\arrayrulewidth}
\end{tabular}
\label{tab:KG statistics}
\end{table}

\section{Hyper-Parameters of the Experiments}
\label{appendix:hyperparametersoftheexperiments}
Table \ref{tab: hyperparameterslinkprediction} and Table \ref{tab:hyperparameterstimeprediction} lists the hyperparameters which we have used for both the datasets for link and time prediction respectively while answering the research questions $1$ and $2$ in Section \ref{sec: experiments}. For both the datasets, while combining with the gadgets, we multiply a coefficient Eta to the overall gadget scores obtained from recurrent and pairwise relation gadgets. 

\begin{table}[H] 
\centering
\small
\caption{Table shows the hyperparameter settings for time prediction over two TKG datasets. LR represents the learning rate used in the training}
\vspace{0.03in}
\begin{tabular}{|c|c|c|}
\Xhline{3\arrayrulewidth}
hyperparameter & YAGO11K & WIKIDATA12k \\
\hline
LR (base) & 1e-3 & 1e-3 \\
LR (w/ gadgets) & 1e-3 & 1e-3 \\
LR (w/ KGE ) & 1e-3 & 1e-3 \\
Eta (w/ KGE) &1 & 0.1 \\
Eta (w /gadgets) & 1 & 0.1 \\
\Xhline{3\arrayrulewidth}
\end{tabular}
\label{tab:hyperparameterstimeprediction}
\end{table}

\begin{table}[H] 
\centering
\small
\caption{Table shows the hyperparameter settings for link prediction over two temporal KG datasets. LR represents Learning Rate used during the training}
\vspace{0.03in}
\begin{tabular}{|c|c|c|}
\Xhline{3\arrayrulewidth}

hyperparameter & YAGO11K & WIKIDATA12k \\
\hline
 LR (base) &  1e-3  & 1e-3 \\
 LR (w/ gadgets) & 1e-3 & 1e-2 \\
 LR (w/ KGE ) & 1e-3 & 1e-2 \\
 Eta (w/ KGE) &  1e-3 & 1e-3 \\
 Eta (w /gadgets)  & 1e-3  & 1e-3 \\
\Xhline{3\arrayrulewidth}
\end{tabular}
\label{tab: hyperparameterslinkprediction}
\end{table}
\section{Implementation Details}
\label{appendix:implementationdetails}
During the process of rule extraction in Section (\ref{subsec:TemporalLogicRuleExtraction}), given the quadruple $(s,r_{h},o,T)$ in the rule head, we avoid this quadruple to re-occur in the body of the rule. \\
There are some cases in Time Prediction where we have $0$ groundings for all the instances with respect to both start and end scores. In such cases, for a given relation $r$, instead of predicting an arbitrary interval, we predict $t\_start$ as $mean\_start[r]$, and $t\_end$ as $mean\_start[r] + mean\_offset[r]$. Here, $mean\_start[r]$ and $mean\_offset[r]$ are the average start and the average offset of intervals for the relation $r$, computed in terms of the assigned ids (as explained in Appendix \ref{appendix: distancebetweentimeinstances}).

\section{Human interpretability of rules}
\label{appendix:humaninterpretabilityofrules}
Here, we give some more examples of how our rules provide human interpretable predictions.\\

\noindent \textbf{Query:} (Ammerschwihr, liate, ?, [1920,present])\\
\noindent \textbf{Answer:}  Haut-Rhin \\
\noindent \textbf{Rule grounding gold object:} \\
$liate(E1,E2,T1) \leftarrow MetBy (T1, T2) \wedge liate(E1,E3,T2) \wedge Equals (T2,T3)\land liate^{-1}(E3,E4,T3) \land Meets (T3,T4) \land liate(E4,E2,T4) $
\\
\noindent \textbf{Grounding:} $E1:$ Ammerschwihr, $E2:$ Haut-Rhin, $E3:$ Upper Alsace, $E4$: Soultzmatt, $T1:$ [1920, present], $T2:$ [1871, 1920], $T3:$ [1871, 1920], $T4:$ [1920, present]\\
Here, $liate$ stands for $located\;in \;the \;$ $administrative\; territorial\;entity$ (WIKIDATA12k).\\
The above rule provides an $explanation$ of why $Ammerschwihr$ is located in the administrative entity of $Haut-Rhin$ since $1920$,  by reasoning that $Ammerschwihr$ and $Soultzmatt$ were both a part of the entity $Upper\;Alsace$ for the same time interval just before $1920$, and $Upper\;Alsace$ became a part of the entity $Haut-Rhin$ in the year 1920.\\

\noindent \textbf{Query:} (Turku, country, Russian Empire, ?)\\
\textbf{Answer:} [1809,1917]\\
\textbf{Rule grounding gold start and gold end:}\\
$country(E1,E2,T1) \leftarrow Meets (T1, T2) $ $  \wedge country(E1,E3,T2) \wedge Equals(T2,T3) \wedge country^{-1}(E3,E4,T3) \wedge MetBy(T3,T4) \wedge country(E4,E2,T4)$ \\
\textbf{Grounding:} $E1:$ Turku, $E2:$ Russian Empire, $E3:$ Finland, $E4:$ Mikkeli Province,  $T2:$ [1917, present], $T3:$ [1917, present], $T4:$ [0,1917]\\\\
The above rule explains that $Turku$ is a part of the country $Finland$ since $1917$, and another place $Mikkeli\;Province$ is also a part of $Finland$ since $1917$. Just before this, $Mikkeli\;Province$ was a part of $Russian\;Empire$, so the rule gives us the $temporal\;information$ that $Turku$ was also a part of $Russian\;Empire$ just before being a part of $Finland$.
\end{document}